\documentclass[journal]{IEEEtran}
\usepackage{amsmath,amsfonts}
\usepackage{bbm}
\usepackage{algorithmic}
\usepackage{algorithm}
\usepackage{array}
\usepackage[caption=false,font=normalsize,labelfont=sf,textfont=sf]{subfig}
\usepackage{textcomp}
\usepackage{stfloats}
\usepackage{url}
\usepackage{verbatim}
\usepackage{cite}
\usepackage{url}
\usepackage{hyperref}
\usepackage{arydshln}
\usepackage{booktabs}
\usepackage[flushleft]{threeparttable}
\usepackage{color}
\usepackage{listings}
\usepackage[absolute]{textpos}

\usepackage{multicol, blindtext}
\usepackage{multirow}
\usepackage{makecell}

\newcommand{\tcb}[1]{{\textcolor{blue}{#1}}}

\newcommand{\tcck}[1]{{\textcolor{black}{#1}}}

\ifCLASSINFOpdf
  \usepackage[pdftex]{graphicx}
\else
  \usepackage[dvips]{graphicx}
\fi
\begin{document}
%

\title{
Quantifying Uncertainty In Traffic State Estimation Using Generative Adversarial Networks}

%
%
%

\author{Zhaobin Mo$^{1}$, Yongjie Fu$^{1}$, and Xuan Di$^{1,2}$
\thanks{Accepted by the 25th IEEE International Conference on Intelligent Transportation Systems (IEEE ITSC 2022)}
\thanks{$^{\text{†}}$Zhaobin Mo and Yongjie Fu contributed equally to this work. }
\thanks{$^{\text{‡}}$This work is partially sponsored by NSF CPS-2038984.}
\thanks{$^{1}$Department of Civil Engineering and Engineering Mechanics,
        Columbia University, New York, NY 10027, USA
        {\tt\small \{zm2302,yf2578,xd2187\}@columbia.edu}.}%
\thanks{$^{2}$Center for Smart Cities, Data Science Institute,
Columbia University, New York, NY 10027, USA.}%
}

\begin{textblock}{17}(0.5,0.2)
\noindent\tcb{Published in: \href{https://ieeexplore.ieee.org/abstract/document/9921791} \textit{2022 IEEE 25th International Conference on Intelligent Transportation Systems (ITSC).}\\
Please cite this paper as: \\Zhaobin Mo, Yongjie Fu, and Xuan Di. ``Quantifying Uncertainty In Traffic State Estimation Using Generative Adversarial Networks." \textit{\\2022 IEEE 25th International Conference on Intelligent Transportation Systems (ITSC). IEEE, 2022.}}
\end{textblock}

\maketitle

\begin{abstract}

This paper aims to quantify uncertainty in traffic state estimation (TSE) using the generative adversarial network based physics-informed deep learning (PIDL). The uncertainty of the focus arises from fundamental diagrams, in other words, the mapping from traffic density to velocity. To quantify uncertainty for the TSE problem is to characterize the robustness of predicted traffic states. Since its inception, generative adversarial networks (GAN) has become a popular probabilistic machine learning framework. In this paper, we will inform the GAN based predictions using stochastic traffic flow models and develop a GAN based PIDL framework for TSE, named ``PhysGAN-TSE". By conducting experiments on a real-world dataset, the Next Generation SIMulation (NGSIM) dataset, this method is shown to be more robust for uncertainty quantification than the pure GAN model or pure traffic flow models. Two physics models, the Lighthill-Whitham-Richards (LWR) and the Aw-Rascle-Zhang (ARZ) models, are compared as the physics components for the PhysGAN, and results show that the ARZ-based PhysGAN achieves a better performance than the LWR-based one.  

\end{abstract}


%
\IEEEpeerreviewmaketitle

\section{INTRODUCTION}

Traffic states, represented by traffic velocity or density, can contain uncertainty from various sources, including model errors, 
measurement noise, 
inherent stochasticity in driving behavior, 
and random initial or boundary conditions. 
Imposed with these stochastic terms, traffic states are essentially spatiotemporal (ST) random fields. 
In this paper, we are primarily focused on uncertainty arising from driving behaviors and their resultant effect on traffic states.
We aim to infer the distribution of traffic state random fields using sparse data 
collected from fixed-location sensors, i.e., loop detectors.

Traffic state estimation (TSE),  a classical traffic problem, has long been studied aiming to estimate traffic states using various sensor data \cite{Seo-17}. 
The methods that predict traffic states and characterize the robustness of the prediction can be categorized into model-based and data-driven methods. 
The model-based methods assume that observed data are generated by underlying models and filtering methods can be applied to these models to propagate uncertainty.
The data-driven methods learn distributions of uncertainty directly from data without using a prior assumption of underlying physics. 
The model based methods suffer from limitations such as the non-Gaussian likelihoods and high-dimensional posterior distributions \cite{yang2019adversarial}, 
while the data-driven approaches require a huge amount of data to learn the posterior distributions from high-dimensional random inputs. 
Thus, an integration of model based and data-driven methods has become a rapidly growing research arena, namely, physics-informed deep learning (PIDL) \cite{Raissi-2018a,Shi-AAAI-2021,mo2020physics,shi2021physics}. 

Recent years have seen a growing trend of learning random fields using machine learning models, including but not limited to: 
Gaussian Processes \cite{yuan2021macroscopic}, 
Bayesian neural networks \cite{mcdermott2019bayesian}, 
and most recently, generative models \cite{yang2019adversarial}.
Among the generative models, generative adversarial networks (GAN) have demonstrated its robustness in various applications, from image or video generation \cite{chu2020learning} to uncertainty quantification (UQ) \cite{patel2021gan}. 
For UQ problems, physics-informed GANs (PhysGAN) is becoming increasingly popular to quantify uncertainty in stochastic differential equations \cite{zhang2019quantifying,yang2019adversarial,yang2020physics, daw2021pid}. 
Little research, though, has been documented when randomness inherently arises from human behaviors or inherent dynamics. 
In other words, the existing literature on PhysGAN integrates a deterministic physics model into GANs because it is assumed that the equations that generate data are deterministic and the prediction contains stochasticity due to randomness from initial or boundary conditions. 
In this paper, in contrast, we will assume that the models that generate data are themselves stochastic due to randomness in parameters, and thus, stochastic physics models need to be integrated to capture the randomness from these sources.

The contributions of this paper include:\\ 
(1) Physics-informed GANs are used to quantify uncertainty in TSE. 
Because traffic, constituted by human drivers, is inherently stochastic due to randomness in driving behaviors, stochastic traffic flow models are incorporated into GANs to help regularize the training process;\\ 
(2) Parameters in physics are estimated jointly with traffic state prediction;\\
(3) PhysGAN-TSE is applied to the real-world dataset, Next Generation SIMulation (NGSIM), to demonstrate its predictive accuracy compared to other baselines.

The rest of the paper is organized as follows: 
Section~\ref{sec:related_works} reviews the state-of-the-art of PIDL on TSE. Section~\ref{sec:preliminary} introduces the preliminaries of GAN based UQ.
Section~\ref{sec:method} introduces how we integrate PIDL and GAN into the framework of PhysGAN for TSE problems. 
Section~\ref{sec:numerical} demonstrates how PhysGAN is applied to NGSIM to characterize uncertainty from the real-world data. 
Section \ref{sec:conclusion} concludes our work and projects future research directions in this promising arena. 

\section{Related Work}
\label{sec:related_works}

For studies that use model-based UQ for TSE, a prior assumption is usually made about the distribution of randomness in inputs or traffic states. The existing practice includes two ways: one to add a Brownian motion on the top of the deterministic traffic flow models, leading to Gaussian stochastic traffic flow models \cite{Yibing-2009}; 
and the other to derive intrinsic stochastic traffic flow models with more complex probabilistic distributions \cite{jabari2012stochastic}. 
The former is more amenable to applying standard filtering methods while the latter is more challenging to deal with and requires a large population approximation 
to extract mean and variance before any filtering can be applied.

If the noise in the output traffic states is non-Gaussian and follows complex distributions, the filtering methods could fail.
\tcck{
GANs have been used to learn complex distributions of stochasticity residing in random fields from ST data \cite{zhang2019quantifying}.}
\tcck{
Key issues with GANs include mode collapse and training instability \cite{wiatrak2019stabilizing}.
One solution is to regularize the training of GANs by incorporating physics into the structure of generators (and potentially discriminators), where unlabeled data are generated from generators that are guided by physics-based models, so that the search space of parameters is narrowed down. 
}

\tcck{PhysGAN have been applied to quantify real-world data uncertainty in various domains, which includes flood prediction \cite{qian2019physics}, blood alcohol concentration prediction \cite{oszkinat2022uncertainty}, and porous media flow modeling \cite{siddani2021machine}.}
PhysGANs have not been used to quantify uncertainty in traffic state estimation. 
Traffic, different from physical systems, can exhibit highly nonlinear randomness arising from inherent human driving behavior. 
Thus, it poses new challenges when PhysGANs are applied.


\section{Preliminaries}
\label{sec:preliminary}

Let us start with a partial differential equation (PDE). 
Define location $x\in [0,L]$ and time $t \in [0,T]$ and $L, T\in\mathbb{R}^+$. 
Then the ST domain of interest is a continuous set of points: $D=\{(x,t)|x \in[0,L],t\in[0,T]\}$. 
A PDE defined over the ST domain is written as:
\begin{equation}
\label{equ:PDE_u}
\begin{split}
  {\begin{array}{*{20}l}
    \  \partial_t \mathbf{u}(x,t) + \mathcal{N}_x[\mathbf{u}(x,t); \mathbf{\lambda}] = 0, (x,t)\in D, \ \  \\
   \ \mathcal{B}_x[\mathbf{u}(x,t)] = 0,
\end{array}}  
\end{split}
\end{equation}
where, 
$\mathcal{N}_x(\cdot)$ is the nonlinear differential operator, 
$\mathcal{B}_x(\cdot)$ is the boundary condition operator,
$\mathbf{u}(x,t)$ is the exact solution of the PDE, and $\lambda$ is the physics parameter vector.

Now we will approximate the PDE solution, $\mathbf{u}(x,t)$, by a (deep) neural network (DNN) parametrized by $\theta$, i.e. $ \hat{\mathbf{u}}(x,t) = f_{\theta}(x,t)$, which is called ``physics uninformed neural network (PUNN)." 
If this PUNN is exactly equivalent to the PDE solution, then we have
\begin{equation}
\label{equ:ftheta}
   \partial_t \hat{\mathbf{u}}(x,t) + \mathcal{N}_x[\hat{\mathbf{u}}(x,t);\lambda] = 0, (x,t)\in D.
\end{equation}
Otherwise we define a residual function:  
\begin{equation}
r_{\theta, \lambda}(x,t) = \partial_t \hat{\mathbf{u}}(x,t) + \mathcal{N}_x[\hat{\mathbf{u}}(x,t); {\lambda}]. 
\end{equation}
If PUNN is well trained, the residual needs to be as close to zero as possible. Since NNs are normally trained with discrete data points, below we will define the potentially accessible training data in the context of TSE.

Assume the dataset consists of 
(1) \textit{(labeled) observation data} $O=\{(x_o^{(i)}, t_o^{(i)}; \mathbf{u}_o^{(i)}) | i=1,...,N_o\}$, 
and (2) \textit{(unlabeled) collocation points} $C=\{(x_c^{(j)}, t_c^{(j)})| j=1,...,N_c\}$. $i$ and $j$ are the indexes of observation points and collocation points, respectively. 
The numbers of observed data and collocation states are denoted as $N_o, N_c$, respectively.
Observation data $O$ are limited to the time and locations where traffic sensors are placed. 
In contrast, collocation points $C$ have neither measurement requirements nor location limitations, and thus are controllable. 
The target values associated with collocation points are used to regularize the loss function.
Accordingly, we define a loss function as the weighted average of data discrepancy and physics discrepancy.  
Denote the weight of the data discrepancy in the loss function as $\alpha$. The loss function is:
\begin{equation}
\label{equ:L_ftheta}
\begin{split}
 {\begin{array}{*{20}l}
\mathcal{L}({\theta, \lambda}) 
&= \alpha \cdot \frac{1}{N_o} \sum_{i=1}^{N_o} \underbrace{\left| \left| \hat{\mathbf{u}}_o(x_o^{(i)},t_o^{(i)})-\mathbf{u}_o^{(i)}\right|\right| ^2}_{\text {data discrepancy}} \\ 
&+ (1-\alpha)  \frac{1}{N_c} \sum_{j=1}^{N_c} 
\underbrace{\left| \left| r_{\theta, \lambda}(x_c^{(j)}, t_c^{(j)}) \right| \right|^2}_{\text {physics discrepancy}}.
\end{array}} 
\end{split}
\end{equation}

We will introduce stochasticity into the PDE solution. 
Below we will discuss how the probabilistic physics-informed neural networks \cite{yang2019adversarial} can be used for UQ. When considering the stochasticity of a PDE, we assume the uncertainty mainly comes from two sources, the intrinsic uncertainty of the physics parameter $\lambda$ and the measurement uncertainty of the variable $\mathbf{u}$. In connection with this, we treat the physics parameter as a random variable, i.e. $\lambda \sim p_{\lambda}(\lambda)$, and assume the variable $\mathbf{u}$ follows a conditional distribution of the spatial-temporal coordinates, i.e. $\mathbf{u} \sim p_{\mathbf{u}}(\mathbf{u}|x,t)$.

After formulating the uncertainty of the physics $\lambda$ and the variable $\mathbf{u}$, a UQ problem further assumes that there exists a mapping between the variable $\mathbf{u}$ and a latent random variabe $\mathbf{z} \sim p_{\mathbf{z}}(\mathbf{z})$, i.e. $G:(x,t,\mathbf{z}) \rightarrow \mathbf{u}$, such that the PDE holds in a stochastic manner: 
\begin{equation}
\label{equ:generator_u}
\begin{aligned}
    \mathbf{u} = G(x,t,  \mathbf{z})  \sim p_{\mathbf{u}}(\mathbf{u}|x,t), 
\mathbf{z} \sim p_{\mathbf{z}}(\mathbf{z}); \lambda \sim p_{\lambda}(\lambda),
\\
 s.t. \mathbb{E}_{p_{\mathbf{z}}(\mathbf{z})p_{\lambda}(\lambda)}\left[ \partial_t \mathbf{u}(x,t) + \mathcal{N}_x \left[\mathbf{u}(x,t); \lambda \right] \right] = 0,
\end{aligned}
\end{equation}
where $\mathbf{z}$ encodes possible randomness sources that lead to stochasticity in the observable output $\mathbf{u}$. 

With Eq.~\ref{equ:generator_u} generating noised output $\mathbf{u}$ that are constrained by a PDE,  
the UQ problem based on adversarial inference is to match 
the conditional distribution of the generated data, denoted as $\hat{\mathbf{u}} = G_{\theta}(x,t,\mathbf{z}) \sim p_{\theta}(\hat{\mathbf{u}}|x,t)$ parametrized by $\theta$, 
and the conditional distribution of the observed output, which is $p_{\mathbf{u}}(\mathbf{u}|x,t)$.
One widely used metric to match two distributions is the reverse Kullback-Leibler (KL) divergence. 

Now we will formulate the UQ problem in the context of conditional GANs. 
The generator is a neural network surrogate model, denoted as $G_{\theta}$, that approximates the ground-truth mapping $G$ mentioned before. The objective of the generator $G_{\theta}$ is to fool an adversarially trained discriminator $D_{\phi}$. 
Then the adversarial UQ problem becomes a min-max game:
\begin{equation}
    \begin{aligned}
    \min _{G_{\theta}} \max _{D_{\phi}} \ 
    \mathcal{L}_{\theta,\phi} = \mathbb{E}_{q(x,t) p_{\mathbf{z}}(\mathbf{z})} \left[ \log D_{\phi}(x,t, G_{\theta}(x,t, z))\right] \\
    +\mathbb{E}_{q(x,t,\mathbf{u})}\left[\log (1-D_{\phi}(x,t,\mathbf{u}))\right],
    \end{aligned}
\end{equation}
where $\theta$ and $\phi$ are the parameters of the generator and the discriminator, respectively.
$q(x,t,\mathbf{u}) = p_{\mathbf{u}}(\mathbf{u}|x,t)q(x,t)$ is the joint distribution of the spatial-temporal coordinates and the output $\mathbf{u}$, 
and $q(x,t)$ is the marginal distribution of the spatial-temporal coordinates. 
$p_{\mathbf{z}}(\mathbf{z})$ is a standard normally distributed random variable. 

With the physics introduced into the generator, the loss functions of the generator $G_{\theta}$ and the discriminator $D_{\phi}$ can be decoupled and defined below, respectively: 
\begin{equation}\label{eqn:loss_G_fig2}
\begin{aligned}
    & \mathcal{L}_{G}(\theta, \lambda) \\
    & = \alpha \mathbb{E}_{q(x_o,t_o)p_{\mathbf{z}}(\mathbf{z})} \left[ D_{\phi}(x_o, t_o,  \hat{\mathbf{u}}_o) \right] 
      + (1-\alpha)  \mathcal{L}_{phy}(\theta, \lambda) \quad \quad \quad \\
    &\simeq \frac{\alpha}{N_o}\sum_{i=1}^{N_o}D_{\phi}(x_o^{(i)},t_o^{(i)}, \hat{\mathbf{u}}_o^{(i)}) +   (1-\alpha) {\cal L}_{phy}(\theta, \lambda),
    \end{aligned}
\end{equation}
\begin{equation}\label{eqn:loss_D_fig2}
    \begin{aligned}
   & \mathcal{L}_{D}(\phi) \\ &= -\mathbb{E}_{q(x_o, t_o)p_{\mathbf{z}}(\mathbf{z})} \left[\log D_{\phi}(x_o, t_o, \hat{\mathbf{u}}_o)\right] \\ 
    &- \mathbb{E}_{q(x_o,t_o, \mathbf{u}_o)} \left[\log(1-D(x_o,t_o, \mathbf{u}_o))\right] \\
    &\simeq -\frac{1}{N_o}\sum_{i=1}^{N_o}
     \log D_{\phi}(x_o^{(i)},t_o^{(i)}, \hat{\mathbf{u}}_o^{(i)}) +\log(1-D_{\phi}(x_o^{(i)},t_o^{(i)}, \mathbf{u}_o^{(i)}))  ,
    \end{aligned}
\end{equation}
\begin{equation}
\begin{aligned} \label{eqn:physics}
&\mathcal{L}_{p h y}(\theta, \lambda) =\mathbb{E}_{q(x_c, t_c)} \left[ \mathbb{E}_{p_{\mathbf{z}}(\mathbf{z}) p_{\lambda}(\lambda)} \left|\left| r_{\theta,\lambda} \right|\right|^{2} \right],\quad \quad \quad \quad \quad
\end{aligned}
\end{equation}
where $\hat{\mathbf{u}}_{o}^{(i)} = G_{\theta}(x_{o}^{(i)},t_{o}^{(i)}, \mathbf{z}^{(i)})$ and $\hat{\mathbf{u}}_{c}^{(j)} = G_{\theta}(x_{c}^{(j)},t_{c}^{(j)}, \mathbf{z}^{(j)})$ are predictions of the generator on observation and collocation points, respectively. 
Note that ${\cal L}_D$ only depends on the observation points, ${\cal L}_{phy}$ only depends on the collocation points, and ${\cal L}_G$ depends on both the observation and the collocation points.

\section{GAN based PIDL framework for TSE (PhysGAN-TSE)}
\label{sec:method}

In this section, we will introduce how PhysGANs are introduced to solve the UQ problem with latent variables. 

\subsection{Problem Statement}

Let $\mathcal{N}^{(1)}[\cdot]$ and $\mathcal{N}^{(2)}[\cdot]$ be two general nonlinear differential operators. 
The problem of interest is to quantify uncertainty in the traffic density $\rho(x,t)$ and velocity $u(x,t)$ fields at each point $(x,t)$ in the ST domain $D$, such that the following PDEs of a traffic flow model can be satisfied:

\vspace{-0.22in}
\begin{equation}
\label{equ-1}
\begin{split}
 {\begin{array}{*{20}l}
    \ \partial_t  \rho(x,t) + \mathcal{N}^{(1)}_{x}[\rho(x,t),u(x,t);\lambda,\xi^{(1)}(x,t)] = \omega^{(1)}(x,t), \ \  \\
   \ \partial_t u(x,t) + \mathcal{N}^{(2)}_{x}[\rho(x,t),u(x,t);\lambda, \xi^{(2)}(x,t)] = \omega^{(2)}(x,t),
\end{array}} 
\end{split}
\end{equation}
\noindent where 
$\left\{\rho(x,t),u(x,t)\right\}$ is the exact solution of PDEs. 
$\xi^{(i)}(x,t)$ and $\omega^{(i)}(x,t),i \in [1,2],$ are potential intrinsic and observation uncertainties, respectively, represented by random fields. 

We aim to develope PUNNs to approximate $\rho(x,t)$ and $u(x,t)$ with time $t$ and location $x$ as inputs, respectively. 
We denote the approximation from PUNN as $\hat{\rho}(x,t)$ and $\hat{u}(x,t)$. 
To use physics to guide the training of PUNN, we customize the residuals on the collocation points as:
\begin{subequations}
\label{equ:residual}
\begin{align}
     r^{(1)}_{\theta, \lambda}(x_c,t_c)= \partial_t \hat{\rho}_c(x_c,t_c) + \mathcal{N}^{(1)}_x[\hat{\rho}_c(x_c,t_c),\hat{u}_c(x_c,t_c);\lambda], \label{eqn:residual_1}  \\
   r^{(2)}_{\theta, \lambda}(x_c,t_c)= \partial_t \hat{u}_c(x_c,t_c) + \mathcal{N}^{(2)}_x[\hat{\rho}_c(x_c,t_c),\hat{u}_c(x_c,t_c);\lambda], \label{eqn:residual_2}
\end{align}
\end{subequations}
\vspace{-0.12in}

\noindent which are defined according to the traffic flow model in Eq.~\ref{equ-1}. 

\subsection{Architecture Design}
We adopt two types of traffic model as the physics component, the Lighthill-Whitham-Richards (LWR) model ~\cite{Lighthill-1955} and the Aw–Rascle–Zhang (ARZ) model~\cite{Aw-Rascle-2000}, to construct the LWR-based PhysGAN (LWR-PhysGAN) and the ARZ-based PhysGAN (ARZ-PhysGAN).

\hfill \break
\textbf{LWR-PhysGAN}. The LWR model is depicted as:
\begin{subequations}
\label{eqn:lwr}
\begin{align}
      \partial_t \rho + \partial_x (\rho u) = 0, \label{eqn:lwr_1}  \\
    u = u_{max} (1 - \rho /  \rho_{max}). 
    \label{eqn:lwr_2}
\end{align}
\end{subequations}
 Eq.~\ref{eqn:lwr_1} is the conservation law; Eq.~\ref{eqn:lwr_2} defines the relation between the traffic density $\rho$ and the traffic velocity $u$, where $\rho_{max}$ and $u_{max}$ are the maximum traffic density and the maximum traffic velocity, respectively. Note that the time $t$ and locations $x$ of $\rho$ and $u$ are omitted for simplicity. Applying Eq.~\ref{eqn:lwr_1} on the collocation points, we can rewrite  Eq.~\ref{eqn:residual_1} as below below:
 \begin{equation}
     \begin{aligned}
     r_{\theta, \lambda}^{(1)} = \partial_t \hat{\rho}_c + \partial_x ( \hat{\rho}_c \hat{u}_c ).
     \end{aligned}
 \end{equation}
With this residual, we can re-write Eq.~\ref{eqn:physics} in the form of the LWR model as:
\begin{equation}
\begin{aligned} \label{eqn:lwr_residual}
\mathcal{L}_{p h y}^{LWR}(\theta, \lambda) &=\mathbb{E}_{q(x_c, t_c)} \left[ \mathbb{E}_{p_{\mathbf{z}}(\mathbf{z})p_{\lambda}(\lambda)}  \left| \left| r_{\theta, \lambda}^{(1)} \right| \right|^{2}  \right].
\end{aligned}
\end{equation}

\begin{figure}[h]
	\centering
	\includegraphics[scale=.21]{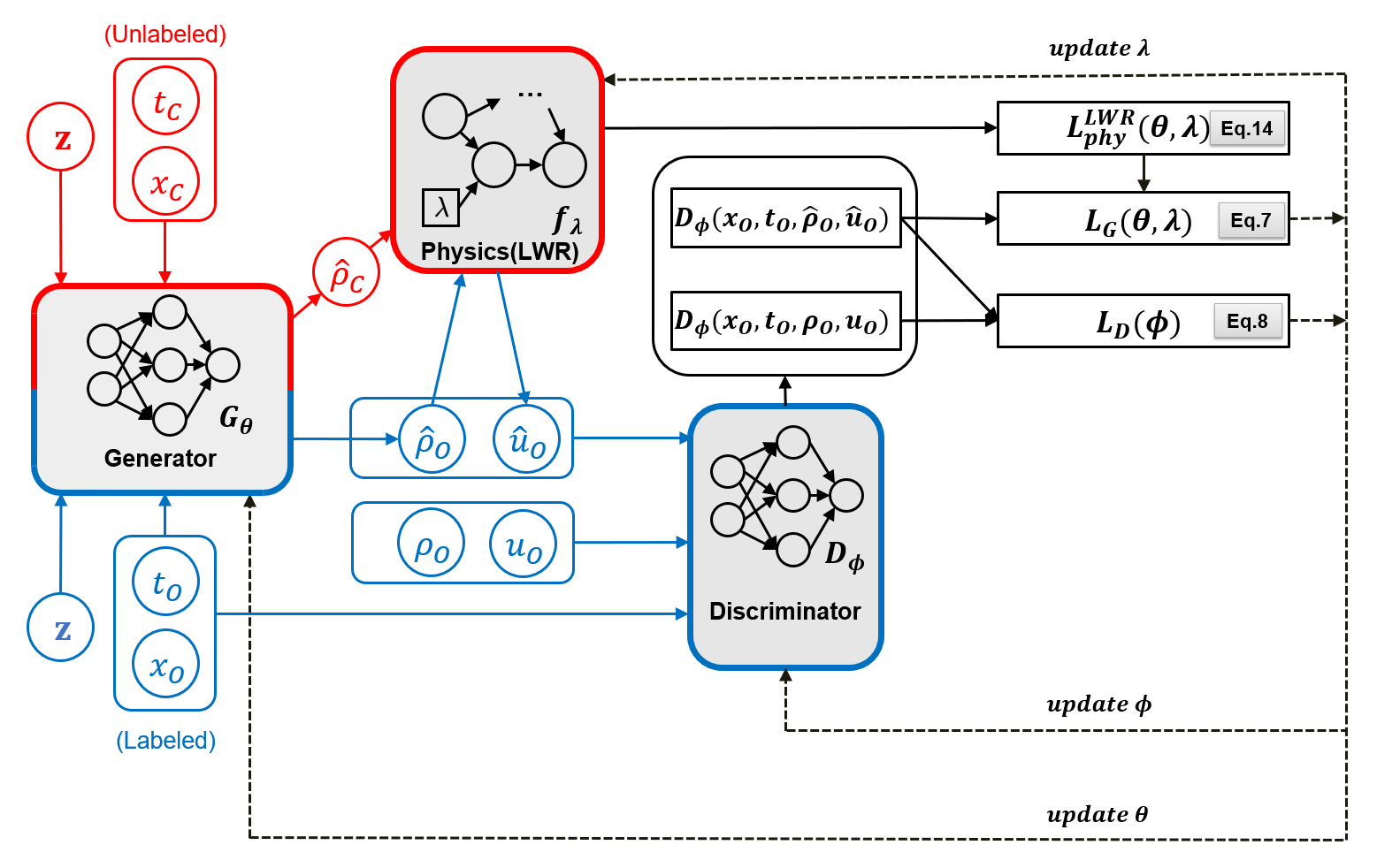}
	\centering 
	\caption{Architecture of the LWR-PhysGAN.}
	\label{fig:lwr}
\end{figure}

The architecture of the LWR-PhysGAN is illustrated in Fig.~\ref{fig:lwr}. It consists of three sub-models, namely the generator $G_\theta$, the discriminator $D_\phi$, and the physics $f_{\lambda}$, which are emphasized by shaded boxes. The blue and red colors illustrate the processes of calculating the data discrepancy and physics discrepancy, respectively. The dashed lines indicate the process of updating parameters.

As for calculating the data discrepancy, both the observation points $(x_o, t_o)$ and the collocation points $(x_c, t_c)$ are fed into the generator $G_{\theta}$ along with the random number $\mathbf{z}$. The outputs of the generator $G_{\theta}$ are the predicted traffic density $\hat{\rho}_o$ and $\hat{\rho}_c$, which correspond to the observation and collocation points, respectively. Then the physics $f_{\lambda}$ takes as input the predicted traffic density $\hat{\rho}_o$ and outputs the predicted traffic velocity $\hat{u}_{o}$ according to the LWR equation depicted in Eq.~\ref{eqn:lwr_2}. The prediction tuple $(x_o, t_o, \hat{\rho}_o, \hat{u}_o)$ and the ground-truth tuple $(x_o, t_o, \rho_o, u_o)$ are then fed into the discriminator $D_{\phi}$ to calculate discriminator loss function $\mathcal{L}_D(\phi)$. 

As for calculating the physics discrepancy, $\hat{\rho}_c$ is used to calculate the physics loss function $\mathcal{L}^{LWR}_{phy}(\theta, \lambda)$ using Eq.~\ref{eqn:lwr_residual}. The physics loss is then added to the generator loss function as a regularization term.

As for updating the parameters, the discriminator loss function $\mathcal{L}_D(\phi)$ is used to update the discriminator parameters ${\phi}$. The generator loss function $\mathcal{L}_{G}(\theta, \lambda)$, which is regularized by the physics loss, is used to update both the generator parameters $\theta$ and the physics parameter $\lambda$.

\begin{figure}[b]
	\centering
	\includegraphics[scale=.22]{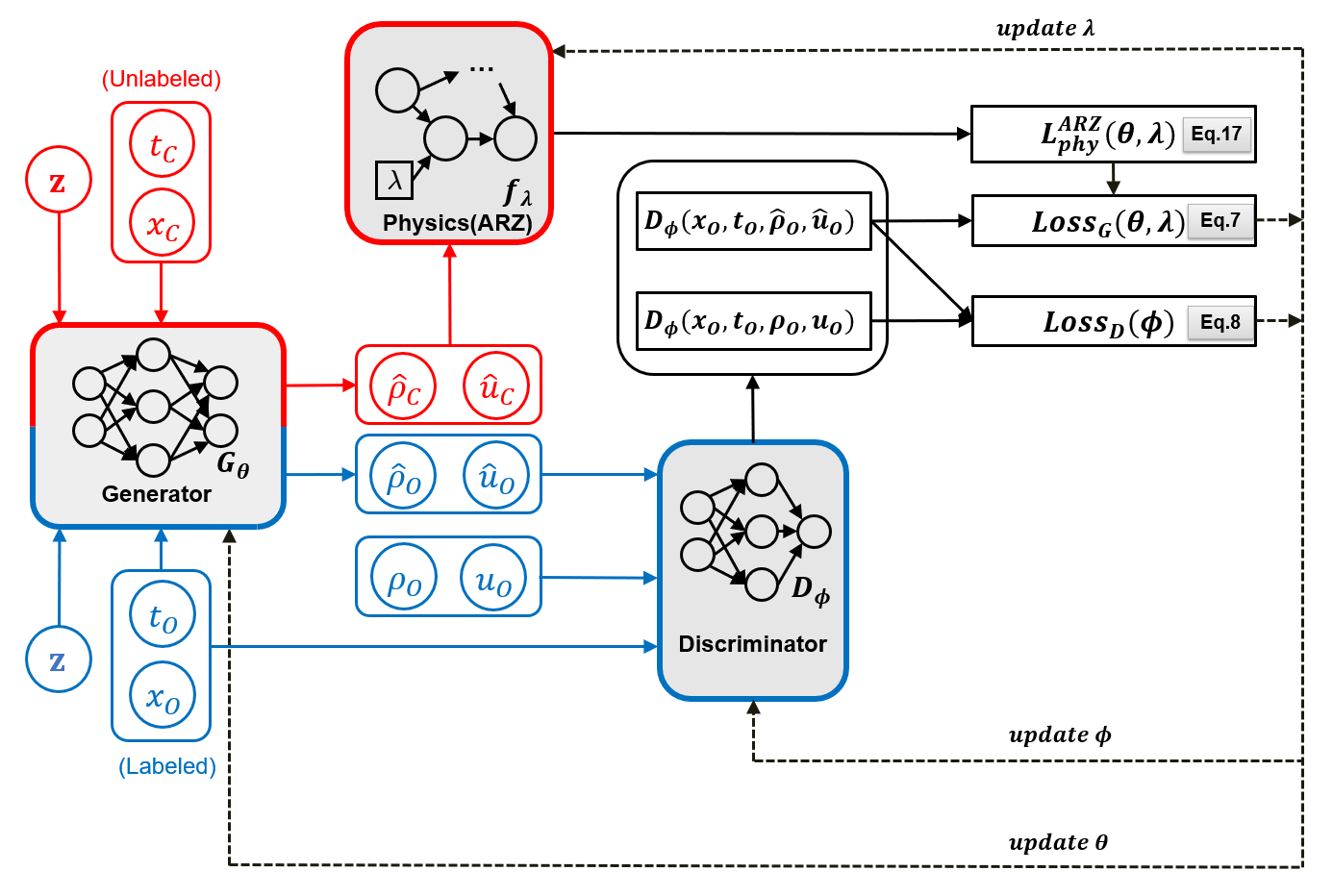}
	\centering 
	\caption{Architecture of the ARZ-PhysGAN.}
	\label{fig:arz}
\end{figure}

\hfill \break
\textbf{ARZ-PhysGAN}. The ARZ model is depicted as:

\begin{subequations}
\label{eqn:arz}
\begin{align}
\partial_t \rho + \partial_x (\rho u) = 0, \label{eqn:arz_1}\\
\partial_t (u+h(\rho)) + u \cdot \partial_x (u+h(\rho))  = (U_{eq}(\rho) - u)/\tau,  \label{eqn:arz_2} \\
h(\rho)=U_{eq}(0)- U_{eq}(\rho),  \\
U_{eq}(\rho)=  u_{max}(1-\rho/\rho_{max})  .
\end{align}
\end{subequations}
Eq.~\ref{eqn:arz_1} is the conservation law, which is the same as Eq.~\ref{eqn:lwr_1}; $\rho_{max}$ and $u_{max}$ are the maximum traffic density and the maximum traffic velocity, respectively; $\tau$ is the relaxation time. Applying Eq.~\ref{eqn:arz_2} on the collocation points, we can rewrite Eq.~\ref{eqn:residual_2} as below:
\begin{equation}
r_{\theta,\lambda}^{(2)} =  \partial_t (\hat{u}_c+h(\hat{\rho}_c)) + \hat{u}_c \cdot \partial_x (\hat{u}_c+h(\hat{\rho}_c))  - (U_{eq}(\hat{\rho}_c) - \hat{u}_c)/\tau.
\end{equation}
Thus, we can re-write Eq.~\ref{eqn:physics} in the form of the ARZ model as:

\begin{equation}
\begin{aligned} \label{eqn:arz_residual}
\mathcal{L}_{p h y}^{ARZ}(\theta, \lambda)
&=\mathbb{E}_{q(x_c, t_c)} \left[ \mathbb{E}_{ p_{\mathbf{z}}(\mathbf{z}) p_{\lambda}(\lambda)} \left| \left| r_{\theta, \lambda}^{(1)}\right| \right|^{2} \right] 
\\
&+ \mathbb{E}_{q(x_c, t_c)} \left[ \mathbb{E}_{p_{\mathbf{z}}(\mathbf{z}) p_{\lambda}(\lambda)} 
 \left| \left|  
r_{\theta, \lambda}^{(2)}
\right| \right|^2 \right].
\end{aligned}
\end{equation}
The structure of the ARZ-PhysGAN is illustrated in Fig.~\ref{fig:arz}, which is similar to the LWR-PhysGAN except for two modifications due to different physics components used:
\begin{enumerate}
    \item The generator $G_{\theta}$ of ARZ-PhysGAN have two outputs, the predicted traffic density $\hat{\rho}$ and the predicted traffic velocity $\hat{u}$.
    \item To compute the physics loss as shown in Eq.~\ref{eqn:arz_residual} , both $\hat{\rho}_c$ and $\hat{u}_c$ are fed into the physics.
\end{enumerate}

\subsection{Training Algorithm}
We use the Adam \cite{kingma2014adam}, one widely used gradient descent algorithm, to update our models. The training algorithm is shown in Algorithm \ref{alg:physgan_alg}. 
\begin{algorithm}[h]
\caption{PhysGAN-TSE Training Algorithm.}
\label{alg:physgan_alg}
\textbf{Initialization}:\\ 
Initialized physics parameters $\lambda^0$.\\
Initialized networks parameters $\theta^0$, $\phi^0$.\\
Training iterations $Iter$. \\
Batch size $m$. \\
Learning rate $lr$.\\
Weights of loss functions $\alpha$.\\
\textbf{Require}: Adam optimizer. \\
\textbf{Input}: The observation data $\{(x_o^{(i)},t_o^{(i)},\mathbf{u}_o^{(i)})\}_{i=1}^{N_o}$ and collocation points $\{x_c^{(j)},t_c^{(j)}\}_{j=1}^{N_c}$.
\begin{algorithmic}[1] 
\label{alg:train}
\FOR{$k \in \{0,...,Iter\}$}
\STATE Sample batches $\{(x_o^{(i)},t_o^{(i)},\mathbf{u}_o^{(i)})\}_{i=1}^{m}$ and $\{x_c^{(j)},t_c^{(j)}\}_{j=1}^{m}$ from the observation and collocation points, respectively \\
\STATE Calculate ${\cal L}_{D}$ by Eq.~\ref{eqn:loss_D_fig2} \\
\STATE $\phi^{k+1} \leftarrow \phi^{k} -lr \cdot \text{Adam}(\phi^{k},\nabla_{\phi} {\cal L}_D)$ \\
\STATE Calculate ${\cal L}_{phy}$ by Eq.~\ref{eqn:physics} \\
\STATE Calculate ${\cal L}_{G}$ by Eq.~\ref{eqn:loss_G_fig2} \\
\STATE $\theta^{k+1} \leftarrow \theta^{k}-lr\cdot\text{Adam}(\theta^{k},\nabla_{\theta} {\cal L}_{G})$ \\
\STATE $\lambda^{k+1} \leftarrow \lambda^{k}-lr\cdot\text{Adam}(\lambda^{k},\nabla_{\lambda} {\cal L}_{G})$ \\
\ENDFOR

\end{algorithmic}
\end{algorithm}

\section{PhysGAN-TSE on Real World Data}
\label{sec:numerical}
To validate the performance and effectiveness of our proposed model and algorithm, in this section, we will apply LWR-PhysGAN and ARZ-PhysGAN to the real-world dataset. 
\subsection{NGSIM Dataset}
The Next Generation SIMulation (NGSIM) dataset
\cite{punzo2011assessment} is an open dataset that is widely used to evaluate various transportation models, which contains high-resolution vehicular information including the position, velocity, occupied lane, and spacing at every 0.1 seconds. 
We focus on trajectories of automobiles from all five lanes in the mainline.

\subsection{Baseline and Metrics}
Two baselines are used for comparison:
\begin{enumerate}
    \item \textbf{Pure GAN}. Pure GAN shares the same architecture with the PhysGAN except for the physics component. We adopt this baseline to verify the effectiveness of adding physics component.  
    \item \textbf{Extended Kalman Filter (EKF)}. EKF applies a nonlinear version of the Kalman filter and is widely used in nonlinear systems like the TSE. We use the ARZ-based EKF, of which details can be found in \cite{shi2021physics}. 
\end{enumerate}
We adopt the relative error (RE) to measure the difference between the mean of the prediction and that of the ground-truth.
Also, we use the KL divergence to quantify the distributional difference between the prediction and the ground-truth.

\subsection{Experiment Setup}
Experiments are conducted on an AWS cloud workstation with 8 Intel Xeon E5-2686 v4 processors and an NVIDIA V100 Tensor Core GPU with 16 GB memory in Ubuntu 18.04.3. 
The initial value of $\rho_{max}$ and $u_{max}$ are 1.0 $m^{-1}$ and 50.0 $m \cdot s^{-1}$, respectively. The learning rate for the Adam optimizer is $0.0005$, and other configurations are kept as default.

\subsection{Results and Discussion}
\begin{figure}
	\centering
	\includegraphics[width=\columnwidth]{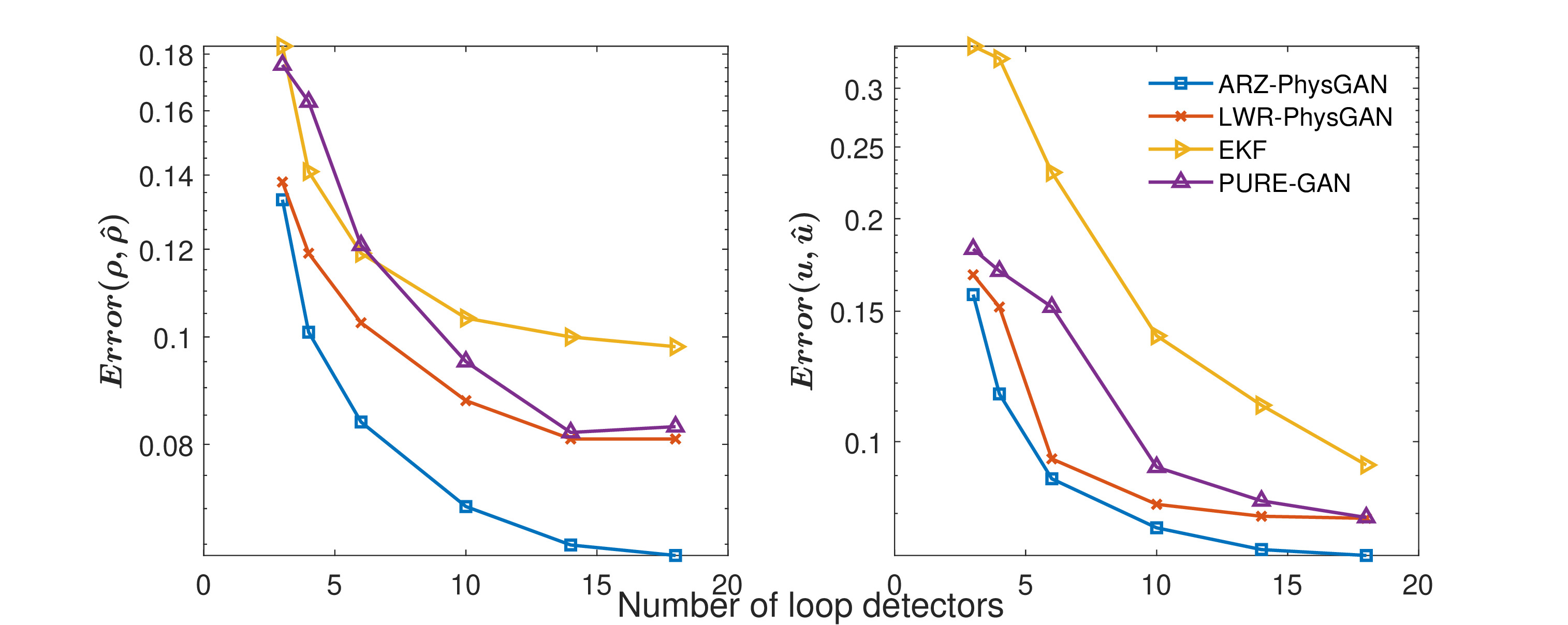}
	\centering 
	\caption{The REs of our proposed PhysGANs and the baselines.}
	\label{fig:graph_rmse}
\end{figure}

\begin{figure}
	\centering
	\includegraphics[width=\columnwidth]{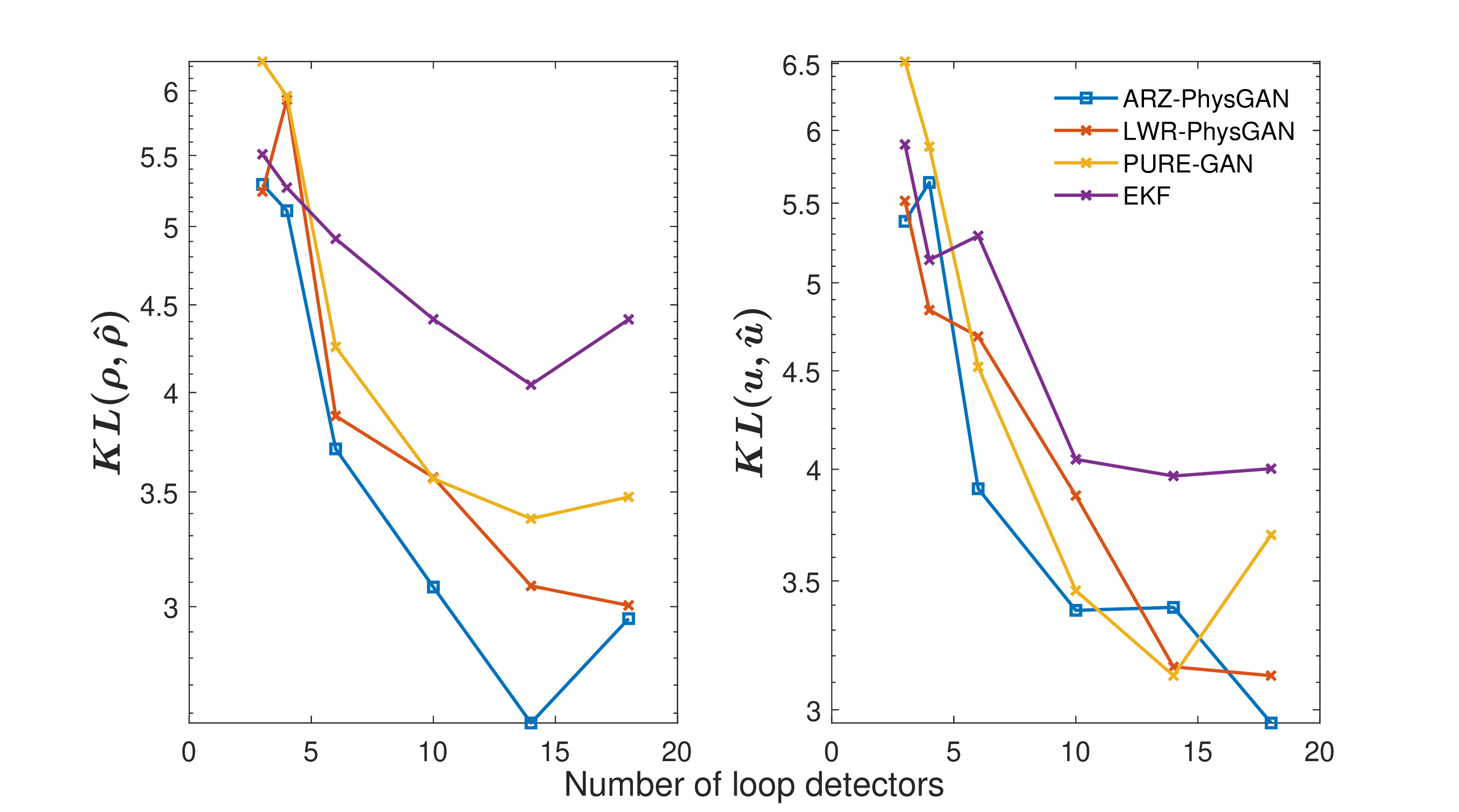}
	\centering 
	\caption{The KL divergence of our proposed PhysGANs and the baselines.}
	\label{fig:graph_kl}
\end{figure}

\textbf{Effect of training sizes}.
Fig.~\ref{fig:graph_rmse} shows the REs of our proposed PhysGANs and the baselines, the left column for the RE of the traffic density and the right column for the RE of the traffic velocity. The x-axis is the number of loop detectors, and the y-axis is the RE. Different scatter types and colors are used to distinguish with different models, namely blue lines with squares for the ARZ-PhysGAN; red lines with crosses for the LWR-PhysGAN; yellow lines with rotated triangles for the EKF; purple lines with triangles for the pure GAN. From this figure, we can see that ARZ-PhysGAN outperforms others across all numbers of loop detectors, followed by LWR-PhysGAN. The superior performance of ARZ-PhysGAN compared to LWR-PhysGAN can be explained by the better compatibility of ARZ with the NGSIM data. In addition, both ARZ- and LWR-PhysGAN outperform the pure GAN by a significant margin, which demonstrates the effectiveness of exploiting physics information. Fig.~\ref{fig:graph_kl} shows the KL divergences under different loop detector numbers. The x-axis is the number of loop detectors, and the y-axis is the value of KL divergence for the prediction of the traffic density (left) and the traffic velocity (right). A lower KL divergence indicates a better performance. Although the LWR-PhysGAN, ARZ-PhysGAN and pure GAN achieve similar KL divergences of the traffic velocity, the ARZ-PhysGAN outperforms other models by a significant margin in terms of the KL divergence of the traffic density. 

\begin{figure}
	\centering
	\includegraphics[width=\columnwidth]{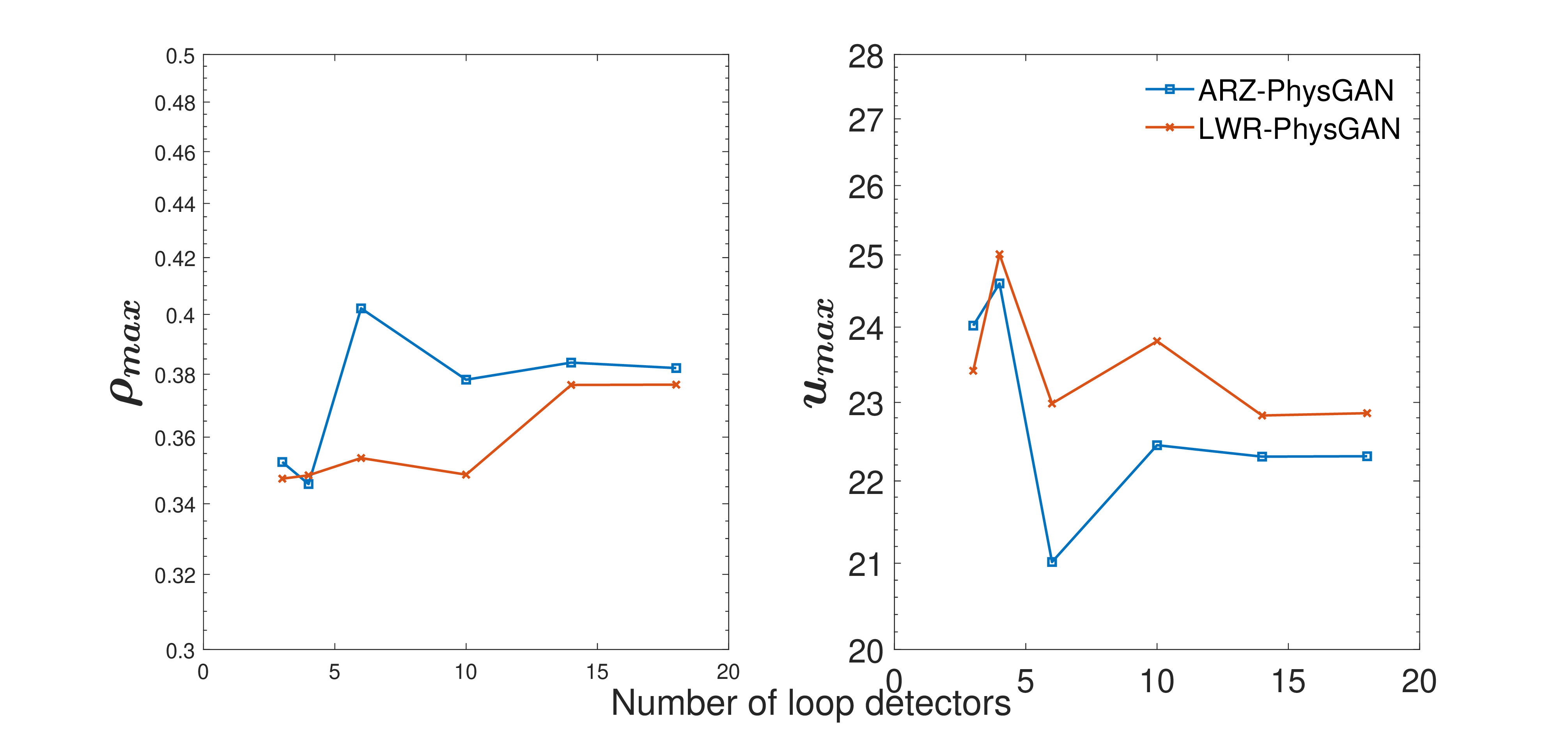}
	\centering 
	\caption{The mean value of the jointly estimated physics parameters for the NGSIM dataset.}
	\label{fig:physpara}
\end{figure}

\textbf{Estimation of physics parameters}. Fig.~\ref{fig:physpara} shows the mean of the estimated physics parameters, the left for $\rho_{max}$ and the right for $u_{max}$, where the x-axis is the number of loop detectors. The line specification is the same as Fig.~\ref{fig:graph_rmse}. From the figure we can see that both $\rho_{max}$ and $u_{max}$ converge as the number of loop detectors increase. The converged values for $(\rho_{max}$,$u_{max})$ of LWR and ARZ model are $(0.3820,22.31)$ and $(0.3766, 22.86)$, respectively, when the number of loop detectors is 18. The units of $\rho_{max}$ and $u_{max}$ are $m^{-1}$ and $m \cdot s^{-1}$, respectively. The converged values of $(\rho_{max}$,$u_{max})$ are reasonable for the highway scenario.

\begin{figure}
	\centering
	\includegraphics[width=\columnwidth]{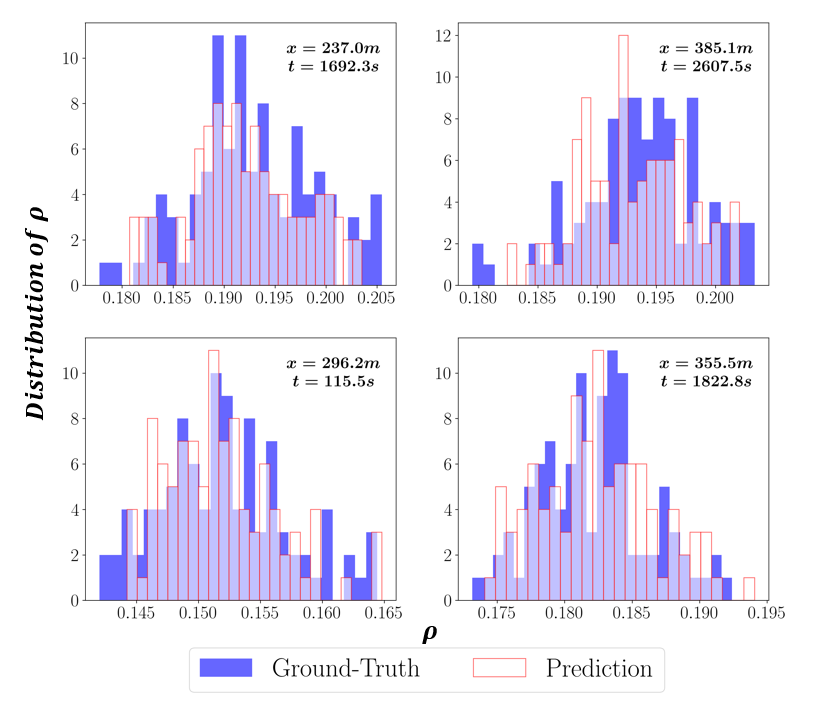}
	\centering 
	\caption{Visualization of predictions of PhysGAN for NGSIM dataset.}
	\label{fig:b1graph}
\end{figure}

\textbf{Visualization of prediction distribution.} Fig.~\ref{fig:b1graph} presents the comparison between the ground-truth traffic density distribution and that predicted by the ARZ-PhysGAN, each subfigure for a randomly sampled spatial-temporal coordinates. The x-axis is the traffic density and the y-axis is the corresponding distribution, where the blue is for the ground-truth and the transparent is for the prediction. Most parts of the predicted and ground-truth distributions overlap with each other, which demonstrates that our proposed model can estimate the real-world traffic states uncertainties well.

\section{Conclusions}\label{sec:conclusion}
This paper proposes PhysGANs to quantify the uncertainty in the TSE problem. The effectiveness of the proposed model is verified by conducting experiments on the real-world NGSIM dataset, where the ARZ-PhysGAN has the best performance, followed by the LWR-PhysGAN. In summary, we show that the GAN model can achieve an enhanced prediction accuracy in terms of the RE and KL divergence if physics information is used to guide the training. We also show that the ARZ-PhysGAN outperforms the LWR-PhysGAN because the ARZ model can better capture the real-world traffic dynamics.

This work can be further improved in two directions. First, although adding physics can help stabilize the training process, GAN is still hard to train due to the min-max game. One possible solution is to use the normalizing flow~\cite{dinh2016density} as the generator, which can explicitly compute the data likelihood and thus the discriminator is not needed. Second, the proposed model needs to be re-trained if applied to other roads or to the same road but within a new time slot, which limits the real-world application of the proposed model. Thus, the model generalizability for unseen scenarios is worth investigating.

\addtolength{\textheight}{-12cm}   




\bibliographystyle{IEEEtran}
\bibliography{ref}

\end{document}